\documentclass[10pt,twocolumn,letterpaper]{article}

\usepackage[pagenumbers]{wacv} 

\usepackage{graphicx}
\usepackage{amsmath}
\usepackage{amssymb}
\usepackage{booktabs}
\usepackage{algorithm}
\usepackage{algpseudocode}
\usepackage{caption}
\usepackage{subcaption}
\usepackage{booktabs}
\usepackage{multirow}
\usepackage{adjustbox}
\usepackage{comment}
\usepackage{rotating}
\usepackage{pifont}
\usepackage[accsupp]{axessibility}

\usepackage[pagebackref,breaklinks,colorlinks]{hyperref}

\usepackage[capitalize]{cleveref}
\crefname{section}{Sec.}{Secs.}
\Crefname{section}{Section}{Sections}
\Crefname{table}{Table}{Tables}
\crefname{table}{Tab.}{Tabs.}

\def\wacvPaperID{494} 
\def\confName{WACV}
\def\confYear{2024}

\begin{document}

\title{Learning Better Keypoints for Multi-Object 6DoF Pose Estimation}

\author{Yangzheng Wu
, Michael Greenspan
\\ RCV Lab, Dept. of Electrical and Computer Engineering, Ingenuity Labs, \\ Queen's University, Kingston, Ontario, Canada \\{\tt \{y.wu, michael.greenspan\}@queensu.ca}
}
\maketitle

\begin{abstract}
We address the problem of keypoint selection, and find that the performance of 6DoF pose estimation methods can be improved when pre-defined keypoint locations are learned, rather than being heuristically selected as has been the standard approach.
We found that accuracy and efficiency can be improved by training a graph network to select a set of disperse keypoints with similarly distributed votes.
These votes, learned by a regression network to accumulate evidence for the keypoint locations, can be regressed more accurately compared to previous heuristic keypoint algorithms.
The proposed KeyGNet, supervised by a combined loss measuring both Wasserstein distance and dispersion, learns the color and geometry features of the target objects to estimate optimal keypoint locations.
Experiments demonstrate the keypoints selected by KeyGNet improved the accuracy for all evaluation metrics of all seven datasets tested, for three keypoint voting methods.
The challenging Occlusion LINEMOD dataset notably improved ADD(S) by $+16.4\%$ on PVN3D,
and all core BOP datasets showed an AR improvement for all objects, of between $+1\%$ and $+21.5\%$. 
There was also a notable increase in performance when transitioning from single object to multiple object training using  KeyGNet keypoints, essentially eliminating the SISO-MIMO gap for Occlusion LINEMOD.
\end{abstract}
\begin{figure}[ht]
    \centering
    \includegraphics[width=0.4\textwidth]{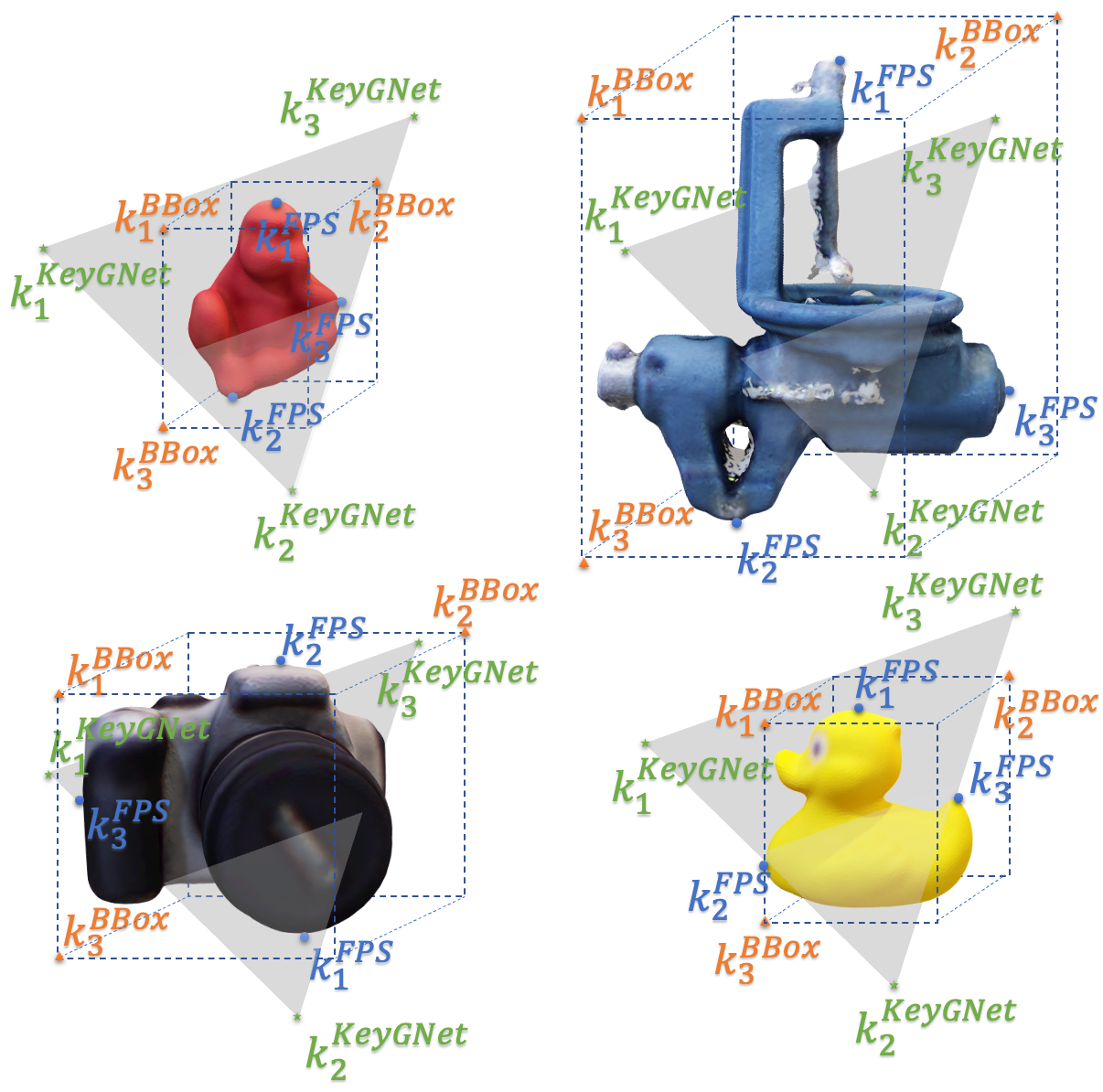}
    \caption{Keypoints sampled by FPS (blue dot), BBox (orange triangle), and KeyGNet (green star). FPS and BBox keypoints are based on the geometry of single objects, while KeyGNet keypoints are learned from the color and geometry of all dataset objects.
    \label{fig:teaser}}
\end{figure}
\section{Introduction}
\label{sec:Introduction}

Estimating the pose of objects in a scene is a fundamental problem in computer vision~\cite{hinterstoisser2012model,posecnn,lin2014microsoft}, which enables a number of important applications such as robot grasping operations~\cite{kleeberger2019large} and augmented reality~\cite{hinterstoisser2012model}. The most basic formulation,
called Six Degree-of-Freedom Pose Estimation (6DoF PE), 
recovers the 3DoF translation and 3DoF rotation parameters of an object that has undergone a rigid transformation. The research community has expanded its efforts to address more general variations, such as allowing deformable transformations~\cite{kumar2016multi,kumar2022organic} and recent one shot training scenarios~\cite{shugurov2022osop,sun2022onepose,chen2023texpose}
which estimate poses without groundtruth values on novel (never-seen and unknown) objects,
albeit with a performance reduction.
Nevertheless,
6DoF PE remains an active area of investigation, with the efficient  pose estimation of multiple objects being a particular focus recently~\cite{sundermeyer2023bop}.

 There are two main approaches to solve 6DoF PE. In the first, the pose is directly regressed by the network~\cite{posecnn,rad2017bb8,kehl2017ssd,chen2021fs,di2022gpv,chen2020g2l}, with the network's output represented as either rotation angles and translation offsets~\cite{kehl2017ssd}, a transformation matrix~\cite{zheng2023hs}, or quaternions~\cite{posecnn}.
 Regression methods are relatively efficient and often implemented as end-to-end trainable architectures~\cite{posecnn,kehl2017ssd,rad2017bb8}.
 While early versions of the direct regression approach tended to lack accuracy~\cite{posecnn}, this has been improved upon recently~\cite{zheng2023hs}.  

\begin{figure*}[ht]
     \centering
        \begin{subfigure}[b]{0.32\textwidth}
         \centering
         \includegraphics[width=\textwidth]{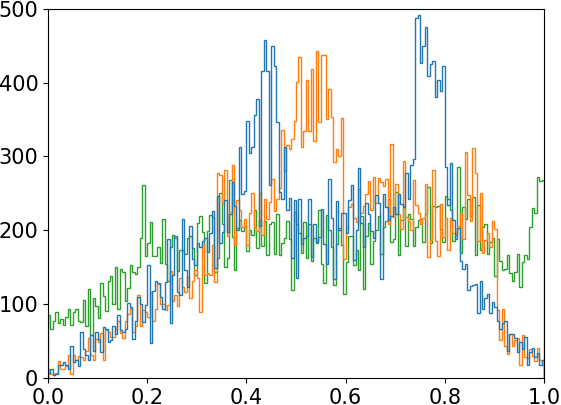}
         \caption{FPS, Wass.  dist.=19.4M }
         \label{fig:fps keypoints}
     \end{subfigure}
     \hfill
     \begin{subfigure}[b]{0.32\textwidth}
         \centering
         \includegraphics[width=\textwidth]{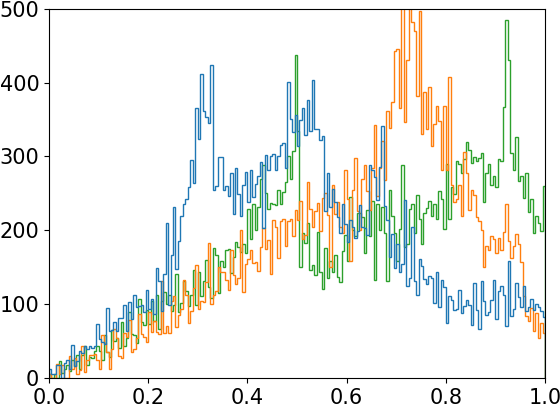}
         \caption{BBox, Wass. dist.=36.8M}
         \label{fig:dispersed keypoints}
     \end{subfigure}
    \hfill
     \begin{subfigure}[b]{0.32\textwidth}
         \centering
         \includegraphics[width=\textwidth]{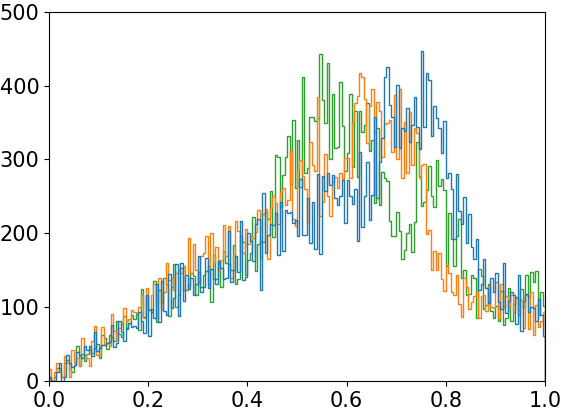}
         \caption{BBox, Wass. dist.=6.4M}
         \label{fig:dispersed keypoints with minimum Wasserstein distance}
     \end{subfigure}
     \caption{
     Histograms of Wasserstein distances to LINEMOD driller model surface points, for 3 sets of 3 keypoints: a) FPS keypoints, b)  BBox keypoints with maximum Wasserstein distance, c) BBox keypoints with minimum Wasserstein distance. ~\label{fig:dispersedvsoptimized}}
\end{figure*}
  
 The second approach 
 are  keypoint-based methods~\cite{he2021ffb6d,pvn3d,pvnet,wu2021vote,rcvpose3d},
 which are the main alternative to direct regression methods. 
 For the 6DoF PE problem, keypoints are defined as a set of 3D coordinates, expressed within an object-centric coordinate reference frame.
Keypoint-based methods are
typically two-stage processes, starting with a keypoint regression network that first estimates keypoint locations within an image. 
The second stage 
then uses model-to-image keypoint correspondences
to estimate pose, often making use of 
classical methods
such as RANSAC or voting-based approaches to increase robustness~\cite{besl1992icp,horn1988closed,fischler1981ransac}.
Keypoint methods have been shown to be among the most accurate solutions to  6DoF PE, with recent works continuing to evolve and advance this approach~\cite{yang2023object,chen2022epro,he2022fs6d,zhou2023deep}.

Despite the success of keypoint methods, the selection of the pre-defined object-centric keypoint locations 
has been overwhelmingly based on 
just two approaches, the first of which is
Farthest Point Sampling (FPS)~\cite{pvnet,pvn3d}.
FPS samples
points
on an object surface based on their relative proximities,
and was originally developed for  progressive image sampling~\cite{FPS97} and subsequently repurposed for keypoint selection~\cite{pvnet}.
The second approach
selects a subset of the corners of
an object
 bounding box (BBox)~\cite{rcvpose3d,wu2021vote}. 
Both of these approaches generate keypoints purely based on the 3D surface geometry of the set of objects,
and ignore other appearance information, such as color.  
Both approaches are also heuristic, with their main objective being to produce keypoints that are geometrically dispersed, and which fall on~\cite{pvnet,pvn3d}, or close to~\cite{wu2021vote,rcvpose3d}, the objects' surfaces.
Their main constraint is that the keypoints are sufficient in number (i.e. $\ge 3$) and distributed in a way (i.e. non-collinearly) so that object transformations can be recovered from the downstream 
model-to-image point correspondences,
e.g. using PnP~\cite{fischler1981ransac} or a suitable alternative~\cite{horn1988closed}.

In this paper, we cast attention to the generation of the pre-defined keypoints themselves. We show that a data-driven generation of the initial object-centric keypoint set can serve to improve the accuracy and the efficiency of existing keypoint-based 6DoF PE methods.
A graph network is trained to optimize a disperse set of keypoints with similarly
distributed votes for keypoint voting
6DoF PE methods~\cite{pvnet,pvn3d,rcvpose3d,wu2021vote}.
The network that encodes the geometry and color information is supervised in a manner that regularizes the learning process, by considering both the distribution
of votes to each keypoint,
as well as the dispersion  
(i.e. geometrical sparseness) 
among them.
Specifically, one loss term considers the Wasserstein similarity of the histograms of voters for each keypoint, and a separate loss term enforces keypoint dispersion.
Some examples of our learned keypoints, along with heuristically generated FPS and BBox keypoints, are shown in Fig.~\ref{fig:teaser}.
Our main contributions are:
\begin{itemize}
    \item We introduce a novel loss function and a graph convolutional network to learn to generate a set of keypoints for a given set of objects.
    \item  We experimentally demonstrate that our generated KeyGNet keypoint sets improve performance of existing keypoint-based 6DoF PE methods. The method not only increases accuracy on networks trained on single objects, but also reduces the performance gap between single and mutliple object scenarios. When using the learned keypoint locations, the training time for the 6DoF PE methods is also reduced.
\end{itemize}
 
To our knowledge, this is the first work that learns the location of pre-defined keypoints for 6DoF PE, rather than generating them heuristically. The main innovation of this work is the concept of learning keypoint locations, and the specific approach to achieve this. The performance improvement is in some cases significant, increasing accuracy by between $1\%$ and $21.5\%$, and reducing training convergence time by between $6h$ and $11h$. 
Our code is available at: \url{https://github.com/aaronWool/keygnet}.

 \subsection{Motivation}
\label{sec:Motivation}
Existing keypoint methods~\cite{pvnet,pvn3d,wu2021vote,rcvpose3d} use regression networks to estimate a quantity
that geometrically relates each image
pixel (and/or point) to each keypoint.
A variety of such quantities have been 
explored in the literature, including 
the offset~\cite{pvn3d},
direction vector~\cite{pvnet},
and radial distance~\cite{wu2021vote,rcvpose3d}
between points.
Once estimated, these quantities are
used to cast votes in an accumulator space,
the collection of which allows for the
robust estimate of the
keypoint locations in the scene.

Fig.~\ref{fig:dispersedvsoptimized} illustrates three histograms of the radial distance quantity~\cite{wu2021vote, rcvpose3d} from each point on the surface of an object's CAD model (the LINEMOD driller), to each of three sets of three keypoints.
Each keypoint set was chosen using a different keypoint selection method.
Each histogram bin represents the number of  votes for that bin value that results using a voting scheme.
Fig.~\ref{fig:fps keypoints} shows the histogram of
this distribution with keypoints
selected using Farthest Point Sampling (FPS).
Fig.~\ref{fig:dispersed keypoints} shows the histogram when the three keypoints are chosen from the eight corners of the bounding box which maximize the Wasserstein distance.
Fig.~\ref{fig:dispersed keypoints with minimum Wasserstein distance} shows the histogram of those three bounding box keypoints with minimal Wasserstein distance.

It can be seen that both FPS
and
maximum Wasserstein BBox keypoints have 
distributions with a greater variance
among the three keypoints,
compared to those selected by 
minimum Wasserstein distance
wherein the distributions of votes are more similar.
As the majority of methods~\cite{pvn3d,pvnet,rcvpose3d} train a single network for all keypoints,
a larger variance between the quantities regressed for each keypoint results in a scenario similar to class imbalance~\cite{he2009learning} in a classification network.
In contrast, 
a reduced variance of the regression quantities, as in Fig.~\ref{fig:dispersed keypoints with minimum Wasserstein distance},
similar to regularization techniques,
can allow for better convergence of the network resulting in both more accurate estimates and faster training,
as demonstrated in Section~\ref{sec:Experiments}.

We repeated this test, and found that the lower variation for the minimum Wasserstein keypoints occurred consistently for all LINEMOD objects.
This preliminary test motivated us to further explore keypoint selection,
and ultimately develop a network structure to learn keypoints that result in similarly distributed votes.

\section{Related Work}
\label{sec:Related Works}
\noindent\textbf{Keypoint Extraction Methods} identify keypoint locations in a scene,
and have been applied 
to 
a variety of computer vision-related problems such as Simultaneous Localization And Mapping (SLAM)~\cite{gomez2019pl,tang2019gcnv2}, Neural Radiance Field (NeRF)~\cite{deng2022depth}, and Non-Rigid Structure from Motion (NrSFM)~\cite{kumar2016multi,kumar2022organic}.
The main objective of these methods is to improve the effective detection of keypoints for better overall performance.
Some of these methods~\cite{chen2022epro,yang2023object} employ a trainable network to estimate keypoints with confidence scores in order to provide redundancies for the consecutive downstream processes.
Others~\cite{wu2021vote,openpose} apply logical or geometric constraints to the keypoints during training.
More recent methods~\cite{he2021latentkeypointgan,suwajanakorn2018discovery} embed keypoints into the latent space of a network so that these latent keypoints can also be trainable 
making the overall structure end-to-end. 

Whereas in the above-described methods, the keypoints are not fixed with respect to some world coordinate reference frame, keypoints have also been pre-defined at fixed locations for problems such as facial recognition~\cite{colaco2020facial,amos2016openface} and human pose estimation~\cite{openpose,singh2020multi},
for which skeleton joints appear as an obvious choice for keypoint locations
to facilitate modeling human motion.
Keypoint selection for some other problems, however,
such as 6DoF PE, is not as apparent.

Most keypoint optimization methods~\cite{duan2019centernet,barroso2019key} focus on fine-tuning the post-processing ignoring the impact of the keypoint selection process. Very few methods~\cite{sun2022dynamic,bosse2009keypoint} address such issues, and find that the overall performance can be improved by  altering  pre-defined keypoints
selection.

\hspace{0.5\baselineskip}

\noindent\textbf{Keypoint-based 6DoF PE Methods}~\cite{pvnet,pvn3d,wu2021vote,rcvpose3d} exhibit relatively good accuracy compared to viewpoint-based methods~\cite{labbe2020cosypose,wang2020self6d,Park2020NOL} or direct regression~\cite{posecnn,oberweger2018making,kehl2017ssd} methods.
Some keypoint-based methods~\cite{park2019pix2pose,oberweger2018making,pavlakos20176} generate hypotheses of the pre-defined keypoint locations by training a network.
The hypotheses are in the form of probability heat maps, and are often filtered by a mask, estimated by a detection network in order to remove  background pixels.
These masks are typically noisy due to the occlusion of scenes in 6DoF PE, and
performance can heavily rely on the robustness of the final least square fitting algorithms.

Keypoint voting-based methods~\cite{pvnet,pvn3d,he2021ffb6d,wu2021vote,rcvpose3d}, however, can better accommodate noise.
The networks cast votes,
typically one per pixel,
to accumulate evidence for each keypoint.
Voting adds redundancy to the detection mask, and
due to the highly redundant and independent nature of voting, the  
resulting
keypoints 
can be more precise,
leading to better overall pose estimation.

\hspace{0.5\baselineskip}

\noindent\textbf{Distribution Similarity Measures} are widely used in ML to quantify the uncertainty of distributions, inspired by information theory.
There are a variety such metrics including KL Divergence (KL Div), JS Divergence (JS Div), and Wasserstein Distance~\cite{kantorovich1960mathematical}.
Cross-entropy~\cite{mannor2005cross} is widely used for classification tasks.
It simplifies the similarity measure by removing the relative entropy of groundtruth in KL Div, 
which is consistent during training.

WGAN~\cite{arjovsky2017wasserstein} justified that Wasserstein distance can optimize GAN training by transforming 
classification
into a regression problem, so that a linear gradient can be created compared to a traditional GAN.
Unlike KL Div and JS Div,
Wasserstein distance is a symmetric metric.
They also demonstrate that Wasserstein distance can still be measured when two distributions do not overlap, which most other metrics cannot.
This often happens in GAN training, and can occur in keypoint voting-based 6DoF PE when a regression network estimates votes for multiple keypoints.

\section{Method: Keypoint Graph Network}
\label{sec:Method}

\begin{figure}[t]
     \begin{center}
              \includegraphics[width=0.38\textwidth]
{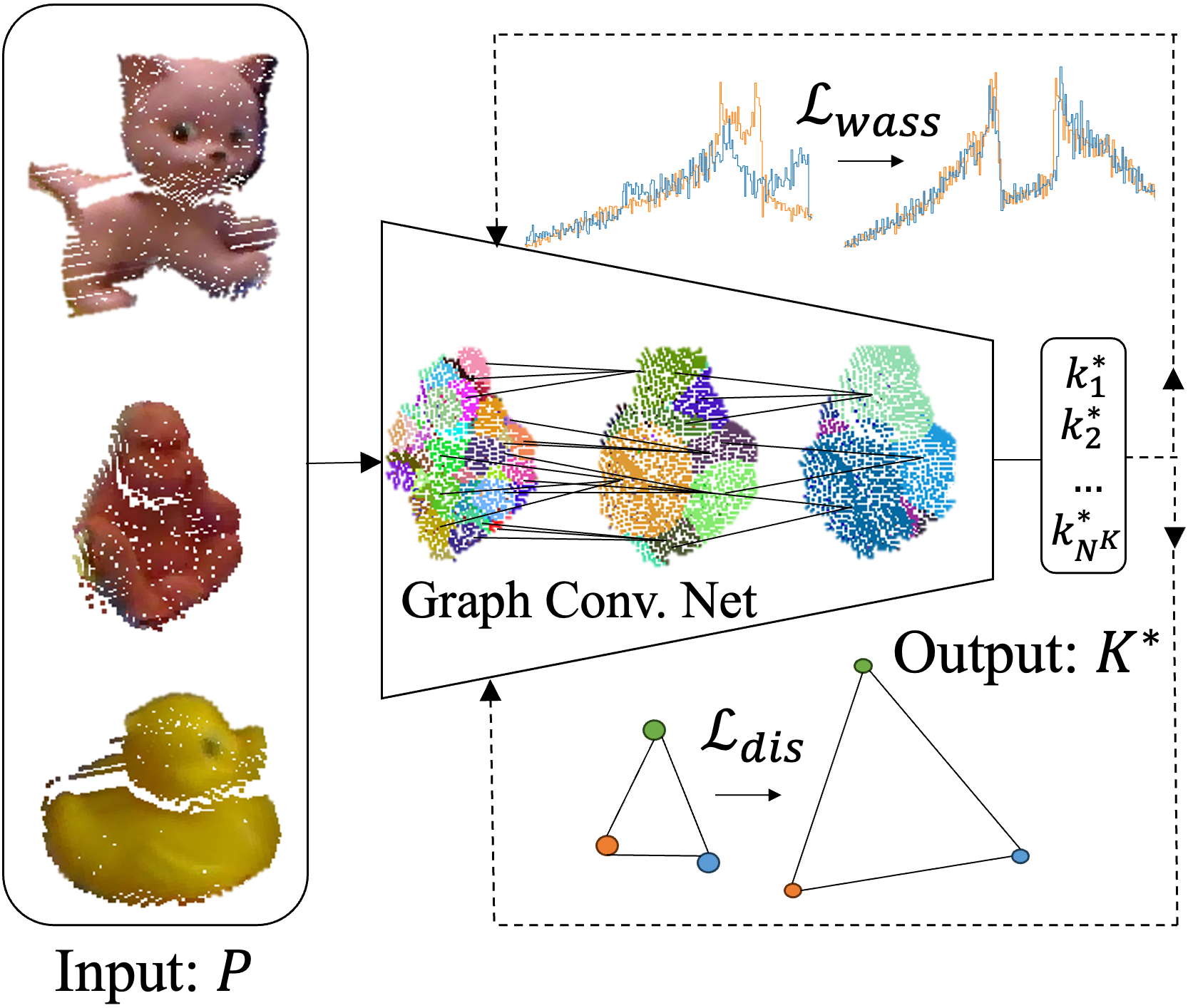}
            \caption{Overview of Keypoint Graph Network (KeyGNet). The network is trained to estimate optimized dispersed keypoint locations with similarly distributed votes for all objects in a dataset.~\label{fig:kgn}}    
     \end{center}
\end{figure}

The preliminary distribution similarity test of keypoint votes motivated us to design a network that learns keypoint locations for a set of objects.
The network, as illustrated in Fig.~\ref{fig:kgn}, is trained to maximize the similarity of the regression quantities between surface points and keypoints.
It is based on a Graph Convolutional Network whose nodes are all points  on the visible (non-occluded) surface of the object of interest in each scene of the training data.

Our loss function comprises two terms,
the first of which enforces the keypoint votes to distribute similarly
for each keypoint.
Wasserstein distance was chosen as the similarty measure, as it has been shown to outperform other popular similarity measures in learned networks~\cite{arjovsky2017wasserstein,gulrajani2017improved}.
The second keypoint dispersion term increases the separation of keypoints, which serves to improve the accuracy of final transformation estimation~\cite{wu2021vote}. 
Our network is trained using all objects in a dataset, resulting in a single set of keypoints for all objects,
which makes the method appropriate for Multiple Instance Multiple Object (MIMO) tasks~\cite{sundermeyer2023bop}.

Let $K\!=\!\left\{k_j\right\}_1^{N_K}$ be a set of keypoints within an object-centric frame, and
let $S\!=\!\left\{p_i\right\}_1^{N_S}$ be a point cloud comprising the non-occluded, visible surface of an object of interest in a scene.  
For each point $p_i$ in $S$, a 
\emph{vote} $v_i^j$ is defined as the regression quantity between $p_i$ and keypoint $k_j$.
For example, if radial voting is used~\cite{wu2021vote}, then vote $v_i^j$ will be a scalar distance, whereas for offset voting~\cite{pvn3d}, $v_i^j$ will be a 3D displacement, etc.

The task is to estimate a keypoint set 
$K^*\!\!=\!\!\left\{k^*_j\right\}_1^{N_K}$
that satisfies two conditions.
The first condition is that
votes $V^j\!=\!\{v_i^j\} \; \forall \; p_i\!\in\!S$ all share a similar distribution:
\begin{equation}
V^j \sim V^i\  
\;\; \forall \;\;
i, j \in [1 \ldots N_K].
\end{equation}
The second condition is that
the keypoints 
$k^*_j$
are geometrically dispersed, so that they lie not too close to each other:
\begin{equation}
\arg \max 
\sum_{i=1}^{N_K} 
\sum_{j=i+1}^{N_K}
|\!|k^*_i - k^*_j|\!|.
\ \end{equation}



Our keypoint graph network (\emph{KeyGNet}) builds on edgeconv~\cite{wang2019dynamic}, in which graphs are dynamically computed in a neighborhood determined using k-nearest neighbors 
within a hierarchy of Voronoi regions of increasing radii.
The convolution kernel operates on the edges 
between points by creating a KNN graph along with a linear operation and a pooling operation.
The edgeconv architecture is unmodified, with the exception of a change to the network classification by reshaping the output from $3\!\times\!C$ classes into $N_K$ normalized keypoints.
At training, the network input is the superset $P$ of all surfaces $S$ in a training dataset, i.e. $P\!=\!\{S_1, S_2 ... S_{N_P}\}$.
The output is the set of optimized keypoints $K^*\!\!=\!\!\{k^*_j\}_1^{N_K}$, which can then be used for both training and inference by any keypoint-based 6DoF PE method.

To supervise  KeyGNet training, we utilize a combined loss with two terms.
The first term is a Wasserstein loss $\mathcal{L}_{wass}$, which is inspired by the work of WassGAN-GP~\cite{gulrajani2017improved} involving both a critic loss and gradient penalty.
The $\mathcal{L}_{wass}$ between two sets of keypoint votes $ V^i$ and $V^j$ is:
\begin{equation}
\begin{split}
    \mathcal{L}_{wass} = \; & \mathbb{E}_{V^i}[D({V^i})] - \mathbb{E}_{V^j}[D({V^j})] \\
    & + \lambda \; \mathbb{E}_{V}[( \left |\!\left| \nabla_{V} D({V}) \right |\!\right| - 1)^2]
\label{eq:W}
\end{split}
\end{equation}

\noindent where $D(V^i)$ and $D(V^j)$ are histogram distributions of votes $V^i$ and $V^j$ for respective keypoints
$k^*_i$
and
$k^*_j$, 
$D(V)$ is the joint distribution of $V^i$ and $V^j$,
$\nabla_{V}$ is a gradient calculated from $V$,
and
$\lambda$ is a gradient penalty hyperparameter.
Instead of applying the gradient penalty only when the generator learns to imitate groundtruth (as in WassGAN-GP~\cite{gulrajani2017improved}), 
we apply it to all votes.
In this way, the distributions of votes are trained to be similar to each other, 
rather than similar to the groundtruth.

The second term is a dispersion loss $\mathcal{L}_{dis}$, which is
inspired by the FPS algorithm:
\begin{equation}
\mathcal{L}_{dis} = e^{-\gamma
\left| \! \left|  
k_i-k_j 
\right| \! \right|
}  
\label{eq:dis}
\end{equation}
where 
$\left| \! \left|  
k_i-k_j 
\right| \! \right|$ 
is the distance separating keypoints $k_i$ and $k_j$.
This term reduces the loss when keypoints stay farther separated, with the value of hyperparameter $\gamma$ chosen so that 
$\mathcal{L}_{dis}\in(0,1]$.
The combined loss will then be:
\begin{equation}
    \mathcal{L} = \alpha\mathcal{L}_{wass} + \beta\mathcal{L}_{dis}
    \label{eq:loss}
\end{equation}
\noindent which is calculated for all pairs of keypoints. The relative values of $\alpha$ and $\beta$ can evolve as training proceeds, to initially confer a greater emphasis on the  $\mathcal{L}_{wass}$ term.

\begin{table*}[ht]
\begin{center}
\begin{adjustbox}{max width=\textwidth}
\begin{tabular}{@{}c|c|l|c|l|c|l|c|l|c|l|c|l@{}}
\hline
\multirow{3}{*}{Dataset}           & \multicolumn{4}{c|}{PVNet}                  & \multicolumn{4}{c|}{PVN3D}                    & \multicolumn{4}{c}{RCVPose}       \\\cline{2-13}
& \multicolumn{2}{c|}{SISO}                  & \multicolumn{2}{c|}{MIMO}   & \multicolumn{2}{c|}{SISO}                  & \multicolumn{2}{c|}{MIMO}    & \multicolumn{2}{c|}{SISO}                  & \multicolumn{2}{c}{MIMO}  \\\cline{2-13}
& FPS & \multicolumn{1}{c|}{KGN} & FPS & \multicolumn{1}{c|}{KGN} & FPS & \multicolumn{1}{c|}{KGN} & FPS & \multicolumn{1}{c|}{KGN} & BBox & \multicolumn{1}{c|}{KGN} & BBox & \multicolumn{1}{c}{KGN} \\\hline \hline
LMO      	&  	61.3	&	64.8 (+3.5)	 &52.7	&64.8 (+12.1)	&	66.4	 &	69.9 (+3.5)	&54.8	&68.7 (\textbf{+13.9})&		73.6	&	76.7 (+3.1)	 &	64.8	&76.4 (+11.6) \\
YCB-V    	&  	77.4	&	79.8 (+2.4)	 &68.2	&78.9 (+10.7)	&	77.8	 &	82.6 (+4.8)	&69.3	&81.9 (\textbf{+12.6})&		85.2	&	88.7 (+3.5)	 &	79.8	&88.2 (+8.4) \\
TLESS    	&  	67.7	&	70.1 (+2.4)	 &62.5	&69.7 (+7.2) 	&	67.3	 &	70.3 (+3)  	&63.2	&70.3 (+7.1)&	    71.5&		75.4 (+3.9)	  &	64.3	&74.9 (\textbf{+10.6}) \\
TUDL     	& 	91.6		&93.1 (+1.5) &	85.7&	93.1 (+7.4) &	90.1     &	93.4 (+3.3) &87.2   &92.9 (+5.7)&       97.8&		98.8 (+1)  	  &  90.2&	98.2 (\textbf{+8}) \\
IC-BIN   	&    70.6&		73.2 (+2.6)	 &65.7	&73.2 (+7.5) 	&	70.4	 &	76.6 (+6.2)	&67.2	&76.1 (\textbf{+8.9})&	    74.0&		76.6 (+2.6)	 &	69.7	&76.2 (+6.5) \\
ITODO    	&  	48.0	&	49.4 (+1.4)	 &27.8	&48.0 (+20.2)	&	49.5	 &	53.9 (+9.8)	&32.4	&53.9 (\textbf{+21.5})&		54.7	&	58.1 (+3.4)	 &	46.5	&58.1 (+11.6) \\
HB       	&    82.5&		85.0 (+2.5)	 &70.9	&84.7 (\textbf{+13.8})	&	82.8	 &	87.6 (+4.8)	&73.5	&87.2 (+13.7)&		87.3	&	89.7 (+2.4)	 &	76.2	&89.7 (+13.5) \\\hline
Average  	&    71.3&	    73.6 (+2.3)	 &61.9	&73.2 (+11.3)	&	72.0	 &	76.3 (+4.3)	&63.9	&75.9 (\textbf{+12})	 &      77  	&	80.6 (+3.6)	  &  70.2&	80.2 (+10) \\\hline
\end{tabular}
\end{adjustbox}
\end{center}
\caption{
Comparison of heuristic (FPS, BBox) vs. KeyGNet (KGN)
performance evaluated by BOP AR.
Keypoints selected by KeyGNet improve the evaluation metrics for all seven BOP core datasets.\label{tab:lmo+ycb+bop}}
\end{table*}

In summary, we train KeyGNet to learn a set of dispersed keypoints with similarly distributed votes, by adapting the keypoint locations to the objects' surface geometries.
The result is a single set of keypoints defined for all objects within a dataset, which can then be used within keypoint-based 6DoF PE methods.

\section{Experiments}
\label{sec:Experiments}

\subsection{Implementation Details}
\label{sec:Implementation Details}
The proposed KeyGNet is trained to generate a set of optimized keypoints, which we tested on 
a variety of 
state-of-the-art 
PE methods~\cite{wu2021vote,pvnet,pvn3d}.
The input segments of KeyGNet
 are the non-occluded visible surfaces of the objects of interest within a scene.
These segments are colored point clouds, derived from training images by applying the camera intrinsics and groundtruth (GT) object pose values.
Layer normalization instead of batch normalization is used in order to apply the gradient penalty in $\mathcal{L}_{wass}$~\cite{gulrajani2017improved}.
We use the default value of $\lambda = 10$ in Eq.~\ref{eq:W} suggested by WGAN-GP. The value of $\gamma = 0.5$ in Eq.~\ref{eq:dis} is based on the diameter of objects in those datasets.
We train KeyGNet with a batch size of 32 fully paralleled on six RTX6000 GPUs.
The optimization used is SGD, with
the initial learning rate set to $lr\!=\!1e\!-\!3$, decaying by a scale of $0.1$ every 50 epochs.

We test the optimized keypoints on three
keypoint-based 6DoF PE voting methods,
RCVPose~\cite{wu2021vote}, PVNet~\cite{pvnet} and PVN3D~\cite{pvn3d}.
These methods use three distinct types of voting, i.e. radial~\cite{wu2021vote}, vector~\cite{pvnet}, and offset~\cite{pvn3d} respectively.
As each PE method uses a different voting scheme, the $D()$ quantities in the Wasserstein loss of Eq.~\ref{eq:W} will be different for each, leading to distinct keypoint sets  for each method,
even when using the same dataset.
These keypoint sets are trained for each PE method, for all objects in each dataset.

We use publicly available implementations of the above three PE networks, as provided by the authors of the original works. These networks are mostly unaltered, the only change being to the output shape to accommodate the shift from SISO (Single Instance Single Object) to MIMO. 
The SISO structures are converted to MIMO by simply training a single regression network for all keypoints of all dataset objects.
For each training run of each PE method, all factors remained the same including the loss function, hyperparameters, network depth, and number of keypoints.

\begin{figure*}
  \centering
    \includegraphics[width=.95\textwidth]{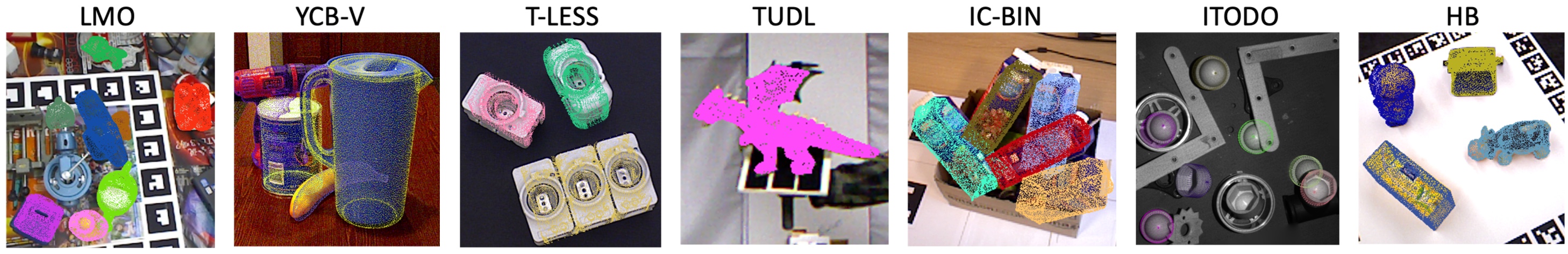}
  \caption{Visualization of poses estimated by KeyGNet keypoints. The overlapped dots are projected point clouds sampled from the object CAD models, applied with the estimated poses.}
  \label{fig:demo}
\end{figure*}
\begin{table}[t]
    \begin{center}
    \begin{adjustbox}{max width=\columnwidth}
    \begin{tabular}{c|c|c|l|l|l}\hline
        PE & \multirow{2}{*}{Mode} & 
        Keypoint & \multicolumn{1}{c|}{LMO} & \multicolumn{1}{c|}{YCB-V} & \multicolumn{1}{c}{Train}\\
        Method& &
         Method & 
        \multicolumn{1}{c|}{ADD(S)} & \multicolumn{1}{c|}{ADD(S) AUC} & \multicolumn{1}{c}{Time}\\\hline
        \multirow{4}{*}{
        PVNet
        }
        
        &  \multirow{2}{*}{SISO} &FPS&40.8&73.4&16h\\\cline{3-6}
        &&KGN&48.0 (+7.2)&79.1 (+5.7)&10h (-6h)\\\cline{2-6}
        &\multirow{2}{*}{MIMO}&FPS&31.7&65.1&22h\\\cline{3-6}
        &&KGN&47.9 (+16.2)&78.1 (\textbf{+13})& 12h (-10h)\\\hline
        \multirow{4}{*}{PVN3D
        }  &  \multirow{2}{*}{SISO} &FPS&63.2&92.3&17h\\\cline{3-6}
        &&KGN&70.5 (+7.3)&95.7 (+3.4) &10h (-7h)\\\cline{2-6}
        &\multirow{2}{*}{MIMO}&FPS&53.9&86.5&23h\\\cline{3-6}
        &&KGN&70.3 (\textbf{+16.4}) & 94.6 (+8.1) & 12h (-10h)\\\hline
        \multirow{4}{*}{RCVPose
        }  &  \multirow{2}{*}{SISO} &BBox&71.1&95.9&16h\\\cline{3-6}
        &&KGN&75.5 (+4.4) & 97.6 (+1.7) & 09h  (-7h)\\\cline{2-6}
        &\multirow{2}{*}{MIMO}&BBox&62&92.5&22h\\\cline{3-6}
        &&KGN&75.6 (+13.6) & 96.6 (+4.1)&11h (\textbf{-11h})\\\hline
    \end{tabular}
    \end{adjustbox}
\end{center}
\caption{
Comparison of heuristic (FPS, BBox) vs. KeyGNet (KGN)
performance evaluated by ADD(S).
KeyGNet improves ADD(S) for both LMO and YCB-V datasets.\label{tab:lmo+ycb}}
\end{table}

\subsection{Datasets and Evaluation Metrics}
\label{sec:Datasets and Evaluation Metrics}
We test our optimized keypoints generated by KeyGNet on 
all Benchmark for 6D Object Pose Estimation (BOP)~\cite{hodavn2020bop} core datasets.
BOP is a benchmark of 6DoF PE providing uniform structures and evaluation metrics for twelve 6DoF PE datasets in various application fields, among which seven are 
considered to be core.
The BOP metrics evaluate the performance on various aspects including visibility, symmetry, and projection when an estimated pose and a GT pose are applied.
The average recall (AR) of Visible Surface Discrepancy (VSD), Maximum Symmetry-Aware Surface Distance (MSSD), and Maximum Symmetry-Aware Projection Distance (MSPD) were tested, 
as well as the overall AR.

We also tested on the original metrics of LINEMOD (LM), Occlusion LINEMOD (LMO)~\cite{hinterstoisser2012model}, and YCB Video (YCB-V) in order to compare against other recent leading non-keypoint-based methods.
ADD(S) was introduced with the LM dataset, and evaluates the average distance (for asymmetric objects)
and the minimum distance (for symmetric objects)
between estimated and GT poses.
The estimate is considered correct when the average distance between the CAD model 
transformed by the estimated and GT poses is within a certain threshold (e.g. $10\%$ of the object diameter).
PoseCNN~\cite{posecnn} proposed ADD(S) AUC, which was designed to be robust by evaluating ADD(S) distance over a series of thresholds.
The accuracy ratio is then plotted in 2D space and connected by a curve.
The overall score is the ratio of the area under the curve versus total area.

\subsection{Results and Time Performance}
\label{sec:Results}
The impact of KeyGNet 
keypoints for the three 6DoF PE methods is shown in Table~\ref{tab:lmo+ycb+bop} and Table~\ref{tab:lmo+ycb},
with some examples shown in Fig.~\ref{fig:demo}.
The heuristic keypoint selection method that led to the best performance was applied to each technique, i.e. FPS for PVNet and PVN3D, and BBox for RCVPose.
In all cases, KeyGNet-selected keypoints boost performance, whether in SISO mode or MIMO mode,
for all three methods and on all seven BOP core datasets.
Specifically, in MIMO mode, the ADD(S) of PVN3D improves by $16.4\%$ on LMO, the ADD(S) AUC of PVNet improves by $13\%$ on YCB-Video, and the BOP AR of PVN3D improves by $12\%$ on average on all seven BOP core datasets. These results indicate that 6DoF PE performance can be impacted and improved, just by altering the location of the pre-defined keypoints. Use of KeyGNet keypoints not only improved the metrics on average over all objects in each dataset, but they also improved the metrics for each individual object (see Supplementary Material Table.S.2, Table.S.3, and Table.S.4).
The observed improvement is likely due primarily to similarly distributed votes being easier to learn by the regression network, compared to other distributions,
as indicated in Fig.~\ref{fig:dispersedvsoptimized}.
A secondary contributing factor is keypoint dispersion, as supported by 
Sec.~\ref{sec:Impact of Weights of Loss Components}. 

The use of KeyGNet keypoints has no impact on inference speed in SISO mode. In MIMO mode, inference speed is improved with a negligible accuracy reduction, as multiple objects share the same KeyGNet keypoints.

\section{Ablation and Tuning Experiments}
\label{sec:Ablation Studies}

\subsection{SISO vs MIMO}
\label{sec:SISO vs MIMO}
BOP introduced the two terms SISO (Single Instance of a Single Object) and MIMO (Multiple Instances of Multiple different Objects) to distinguish between varying levels of challenge in solving 6DoF PE for different types of scenes. 
The majority of previous keypoint-based methods~\cite{wu2021vote,rcvpose3d,pvn3d,pvnet} were designed to address the less challenging SISO case by only processing a single object at a time, which allows the training of unique network parameters for each distinct object.
Motivated in part by the most recent edition of the BOP Challenge competition~\cite{hodavn2020bop}, several recent methods~\cite{he2021ffb6d,li2023multi} address the MIMO case, with multiple object categories trained within a single network.

Networks trained for MIMO are typically less accurate than those trained for SISO,
likely due to a degradation of accuracy of the regression network. 
In contrast, our method can effectively handle the MIMO case,
with little or no degradation in accuracy.
One single set of optimized keypoints returned by KeyGNet for all dataset objects 
serves to simplify the learning process of the subsequent 6DoF PE, reducing the SISO-MIMO performance gap.
\begin{table}[]
\begin{center}
\begin{adjustbox}{max width=\columnwidth}
\begin{tabular}{@{}c|c|c|c|c|c|c@{}}
\hline
LMO & \multicolumn{2}{|c|}{PVNet} & \multicolumn{2}{|c|}{PVN3D} & \multicolumn{2}{|c|}{RCVPose}\\\cline{2-7}
object     & FPS  & KGN & FPS  & KGN & BBox  & KGN   \\\hline
ape         &-7.6&\textbf{-0.3}&-8.2&\textbf{-0.3}& -4.2  & \textbf{-0.4}   \\
can         &-12.2&\textbf{0}&-11.2&\textbf{0}& -2.9    & \textbf{0}     \\
cat         &-7.1&\textbf{0}&-7.6&\textbf{-0.3}& -5.7  & \textbf{-0.2}   \\
driller     &-8.2&\textbf{0}&-10.3&\textbf{0}& -1.3  & \textbf{+0.4}   \\
duck        &-11.3&\textbf{-0.3}&-9.7&\textbf{-0.4}& -7.2    & \textbf{+0.4}   \\
eggbox      &-12.2&\textbf{0}&-10.2&\textbf{0}& -2.1  & \textbf{+0.2}   \\
glue        &-6.7&\textbf{0}&-9.8&\textbf{-0.4}& -6.3  & \textbf{+0.2}   \\
holepuncher &-7.3&\textbf{0}&-7.2&\textbf{0}& -2.1  & \textbf{0}   \\\hline
average     &-9.1&\textbf{-0.1}&-9.3&\textbf{-0.2}& -6.3 & \textbf{+0.1}\\\hline
\end{tabular}
\end{adjustbox}
\caption{SISO-MIMO ADD(S) performance gap on LMO. The change when converting from SISO to MIMO, for heuristic 
and KeyGNet (KGN) keypoint selection. The ADD(S) change 
is small for MIMO training
using 
KGN keypoints.\label{tab:sisoVSmimo(lmo)}}
\end{center}
\end{table}
\begin{table}[ht]
\begin{center}
\begin{tabular}{c|c|c|c|c|c}
\hline
\multicolumn{2}{c|}{PVNet} & \multicolumn{2}{c|}{PVN3D} & \multicolumn{2}{c}{RCVPose} \\ \hline
FPS & KGN & FPS & KGN & BBox & KGN \\ \hline
 -6.3	& \textbf{-1.0}	& -6.7	 &\textbf{-1.2}	&-3.4	&\textbf{-1.0} \\ \hline
\end{tabular}
\caption{SISO-MIMO 
ADD(S) AUC performance gap on YCB-V. The average change 
for all objects, converting from SISO to MIMO, for heuristic 
and KeyGNet (KGN) 
keypoints. The change is small when trained simultaneously on all objects
using 
KGN.
\label{tab:sisoVSmimo_sum(ycb)-simple}
}
\end{center}
\end{table}

To illustrate this, we conduct an experiment comparing the performance change when a single network is first trained individually for each dataset object (SISO), and is then trained simultaneously for all objects (MIMO). In both  scenarios, the network parameters were initialized randomly, from a
standard normal distribution (z-distribution).

The results are shown for each LMO object in Table~\ref{tab:sisoVSmimo(lmo)},
and averaged for all YCB-V objects in
Table~\ref{tab:sisoVSmimo_sum(ycb)-simple}.
(For all YCB-V results, see Supplementary Material
Table S.3.) For the three PE methods, the change in ADD(S) when training for single vs. multiple objects is listed, when using either heuristic (FPS or BBox) or KeyGNet (KGN) keypoints.
The performance gap is significantly reduced when the network is converted from SISO to MIMO.
While degradation existed for all objects and all three PE methods using heuristic keypoints,
no degradation resulted for most LMO objects using KeyGNet keypoints,
with a much smaller average degradation for YCB-V.
Interestingly, RCVPose even demonstrates a small improvement of $+0.1$ on average in LMO.
This experiment again shows that the distribution of votes has a vital impact on the performance of the network, closely correlated to overall 6DoF PE performance.

\subsection{Distribution Similarity Losses of KeyGNet}
\label{sec:Distribution Similarity Losses of KeyGNet}
There are a variety of distribution similarity metrics such as KL Divergence, JS Divergence, and cross entropy, which have been discussed in Sec.~\ref{sec:Related Works}.
The distribution of the comparison of votes in KeyGNet is similar to the generator supervision in WGAN,
which used Wasserstein distance.
We conduct an experiment to compare different distribution similarity losses against $\mathcal{L}_{wass}$ defined in Eq.~\ref{eq:W}.

\begin{table*}[]
\begin{center}
\begin{tabular}{@{}c|c|c|c|c|c|c|c|c|c|c|c|c@{}}
\hline
LMO & \multicolumn{4}{|c|}{PVNet} & \multicolumn{4}{|c|}{PVN3D} & \multicolumn{4}{|c}{RCVPose}\\\cline{2-13}
object     & $\mathcal{L}_{kl}$  & $\mathcal{L}_{js}$ & $\mathcal{L}_{ce}$ & $\mathcal{L}_{wass}$ & $\mathcal{L}_{kl}$  & $\mathcal{L}_{js}$ & $\mathcal{L}_{ce}$& $\mathcal{L}_{wass}$ &$\mathcal{L}_{kl}$  & $\mathcal{L}_{js}$ & $\mathcal{L}_{ce}$ &$\mathcal{L}_{wass}$ \\\hline
ape         	&+20.2	&\underline{+22.0}	&+20.3	&\textbf{+22.6}	&+20.3	&+22.5	&+21.4	&\textbf{+22.9}	&+9.6	    &+9.6	    & +8.6	    & \textbf{+11.7} \\
can         	&+11.8	&\underline{+12.8}	&+12.3	&\textbf{+13.8}	&+11.5	&\underline{+12.6}	&+12.4	&\textbf{+14.1}	&+13.4	    &\underline{+15.4}	& +13.0	& \textbf{+15.6} \\
cat         	&+14.9	&\underline{+15.8}	&+15.1	&\textbf{+16.9}	&\underline{+15.2}	&+14.5	&+14.8	&\textbf{+17.1}	&+12.6	    &\underline{+13.2}	& +13.1	& \textbf{+14.1} \\
driller     	&+13.1	&\underline{+14.2}	&+13.2	&\textbf{+15.0}	&\underline{+13.8}	&+13.2	&+10.2	&\textbf{+15.4}	&+9.0	    &\underline{+10.2}	& +8.0	    & \textbf{+11.1} \\
duck        	&+15.8	&\underline{+17.2}	&+15.9	&\textbf{+18.0}	&+16.1	&\underline{+16.8}	&+16.7	&\textbf{+18.0}	&+21.1	    &\underline{+22.6}	& +21.8	& \textbf{+23.1} \\
eggbox      	&+9.9	&\underline{+10.9}	&+10.0	&\textbf{+11.9}	&+9.1	&+10.0	&\underline{+10.4}	&\textbf{+12.2}	&+9.6	    &\underline{+10.6}	& +8.6	    & \textbf{+11.9} \\
glue        	&+16.1	&\underline{+17.3}	&+16.8	&\textbf{+18.2}	&\underline{+16.6}	&+16.3	&+15.9	&\textbf{+18.2}	&+6.6	    &\underline{+8.2}	    & +5.4     & \textbf{+8.7} \\
holepuncher 	&+10.7	&\underline{+12.5}	&+11.2	&\textbf{+13.3}	&+11.2	&\underline{+12.6}	&+12.0	&\textbf{+13.7}	&+9.7	    &\underline{+11.7}	& +9.1	    & \textbf{+12.7} \\\hline
average     	&+14.1	&\underline{+15.3}	&+14.3	&\textbf{+16.2}	&+14.2	&\underline{+14.8}	&+14.2	&\textbf{+16.4}	&+11.5	    &\underline{+12.7}	& +11.0	& \textbf{+13.6} \\\hline
\end{tabular}
\caption{ADD(S) Improvements of KeyGNet Distribution Similarity Losses tested on LMO in MIMO mode. keypoints selected by $\mathcal{L}_{wass}$ lead to more improvements on all three PE methods compared to the other losses. $\mathcal{L}_{js}$ improves $+1.2\%$ more than $\mathcal{L}_{kl}$ and $\mathcal{L}_{ce}$ on average.\label{tab:losses_lmo}}
\end{center}
\end{table*}

We train KeyGNet by applying losses based on different metrics (KL Div Loss $\mathcal{L}_{kl}$, JS Div Loss $\mathcal{L}_{js}$, Cross Entropy Loss $\mathcal{L}_{ce}$) on LMO, on all three 6DoF PE methods. The results are shown in Table~\ref{tab:losses_lmo}.
$\mathcal{L}_{wass}$ improves the performance the most compared to the other losses,
scoring highest for each individual
object for each PE method. 
The second best performance was $\mathcal{L}_{js}$, followed by the $\mathcal{L}_{kl}$ and $\mathcal{L}_{ce}$ which had similar improvements.
The result of this experiment confirmed the WGAN conclusion, and our assumption made in Sec.~\ref{sec:Related Works} that,
even when the distribution of keypoint votes has no overlap, $\mathcal{L}_{wass}$ can still measure the differences and calculate 
gradients for the network compared to other similarity measures.

\subsection{Number of Keypoints}
\label{sec:No. of Keypoints}
\begin{figure}[t]
     \begin{center}
     \includegraphics[width=0.45\textwidth]{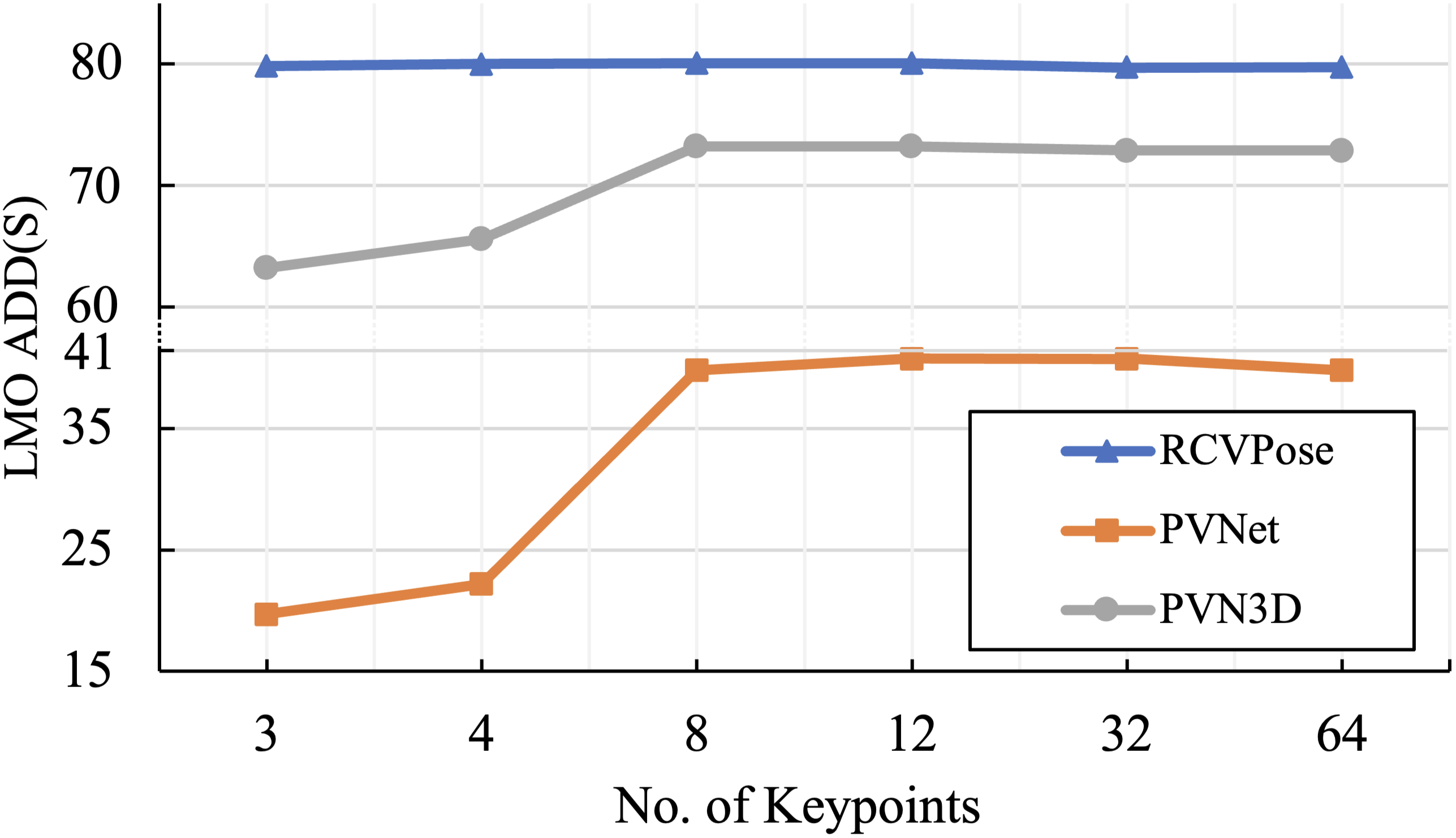}
     
     \caption{ADD(S) vs. number of KeyGNet keypoints, LMO dataset. PVNet
     and PVN3D
     improve for more than 3, and saturate at 8 keypoints. RCVPose saturates at 3 keypoints.
     }
     \label{fig:noofkeypoints}
     \end{center}
\end{figure}
PE methods use various numbers of keypoints, to balance time and accuracy.
Some~\cite{pvn3d,pvnet,wang2021gdr} argue that more keypoints provide redundancy to the least square fitting algorithm ultimately used in the final transformation estimation, whereas others~\cite{wu2021vote,rcvpose3d} use as few as three keypoints to ease the estimation task of the backbone network.

We conduct an experiment testing the impact of the number of KeyGNet keypoints on network accuracy.
Fig.~\ref{fig:noofkeypoints} shows the ADD(S) trend of three PE methods with optimized keypoints training with the optimized keypoints in MIMO mode,
using all LMO objects.
PVNet~\cite{pvnet} and 
PVN3D~\cite{pvn3d}
exhibited a slight overall improvement in ADD(s) as the number of keypoints increased, which saturated when there are more than eight keypoints.
The ADD(S) of RCVPose~\cite{wu2021vote}
stayed fairly constant,
independent of an increase in the number of keypoints beyond three.
In subsequent experiments, we chose the number of keypoints that optimized performance
for each PE method, i.e. 8 keypoints for PVNet and PVN3D, and  
3 for RCVPose.
\subsection{Loss Component Weights}
\label{sec:Impact of Weights of Loss Components}
The KGNet loss function (Eq.~\ref{eq:loss}) comprises two terms.
$\mathcal{L}_{wass}$ supervises the distribution of votes associated with each keypoint whereas $\mathcal{L}_{dis}$ keeps the keypoints dispersed.
These two components can compete with each other simply because those keypoints with a similar distribution of votes tend to cluster within a local neighborhood.
Training with an equal weight of both components can therefore cause the network to converge to a local minimum,
with closely clustered keypoints, which therefore are less effective at 
the ultimate PE goal of
transformation estimation~\cite{wu2021vote}.

\begin{figure}[t]
     \centering
     \includegraphics[width=0.45\textwidth]{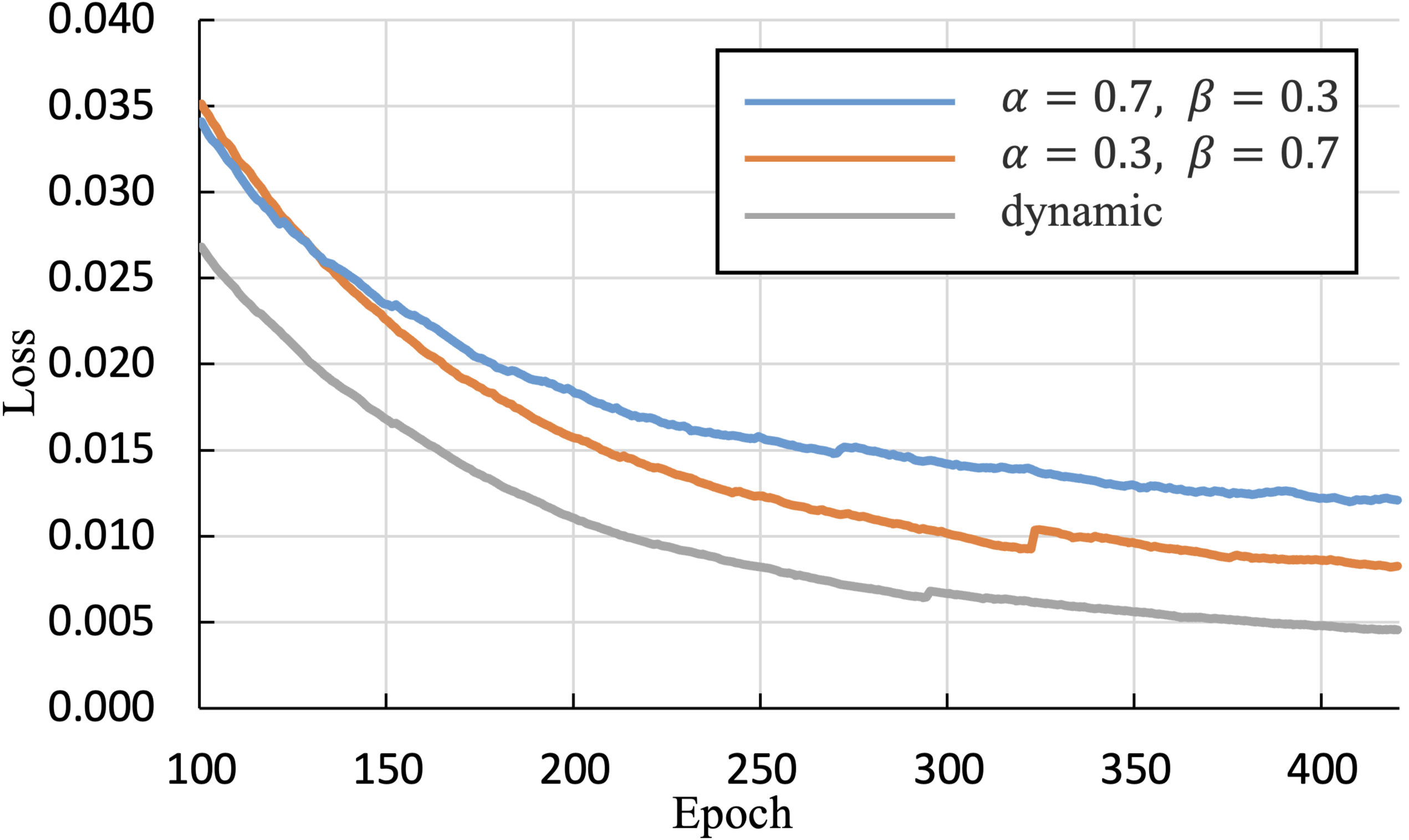}
     \caption{Impact of weighting loss components of Eq.~\ref{eq:loss}. KeyGNet converges either slowly ($\alpha\!\!=\!\!0.7,\beta\!=\!0.3$) or less accurately ($\alpha\!=\!0.3,\beta\!=\!0.7$) with fixed weights. 
     Dynamic weight scheduling improves training by swapping weight values at epoch 50.}
     \label{fig:loss}
\end{figure}
To address this, we experiment with different combinations of 
$\alpha$ and $\beta$, which are the weighting factors of the respective loss terms. The smoothed loss curve 
for training RCVPose on LMO for 3 keypoints
is plotted in Fig.~\ref{fig:loss}.
This plot shows that
KeyGNet converges faster when $\alpha\!=\!0.7$ and $\beta\!=\!0.3$, whereas it is more accurate but converges slower when $\alpha\!\!=\!\!0.3$ and $\beta\!\!=\!\!0.7$.
We therefore train the network
using the plotted dynamic schedule, which weights 
$\mathcal{L}_{wass}$ more heavily 
($\alpha\!=\!0.7$) initially, and shifts to weigh $\mathcal{L}_{dis}$ more heavily
($\beta\!=\!0.7$)
at epoch 50.  This schedule has the benefit of converging both
efficiently and accurately.

\subsection{Initial Input of KeyGNet}
\label{sec: Initial Input of KeyGNet}
\begin{table}[]
\begin{center}
\begin{adjustbox}{max width=\columnwidth}


\begin{tabular}{@{}c|c|c|c@{}}
\cline{2-4}
& \multicolumn{2}{c|}{SISO} & \multicolumn{1}{c}{MIMO} \\
\hline
LMO object           & FPS-KGN & BBox-KGN 
& KGN \\\hline
ape         & \textbf{65} & \underline{64.9} 
& \textbf{65}   \\ 
can         & \underline{96} & \underline{96} 
& \textbf{96.4}   \\ 
cat         & 57.5 & \underline{57.9} 
& \textbf{58}   \\ 
driller     & 79.9 & \underline{80.2} 
& \textbf{82.1}   \\ 
duck        & 64.7 & \underline{65.2} 
& \textbf{65.6}   \\ 
eggbox      & \underline{80.7} & \underline{80.7} 
& \textbf{82.2  }   \\ 
glue        & 73.9 & \underline{74.2} 
& \textbf{75.1}   \\ 
holepuncher & 79.8 & \underline{80.7} 
& \textbf{81.2}   \\ \hline
average     &  74.7  & \underline{75} 
& \textbf{75.6}   \\\hline
\end{tabular}
\end{adjustbox}
\end{center}
\caption{ADD(S) for different initial KeyGNet inputs, tested on LMO using RCVPose.
Segment initialization (KGN) is slightly more accurate than heuristic keypoint initialization.\label{tab:initPoints}}

\end{table}
At training,
KeyGNet accepts as input 
all points on  the visible surface of objects within a scene (segments).
An alternative could input fewer initial keypoints, sampled by FPS or BBox.

To explore the impact of initialization, we conduct an experiment to compare performance with keypoints optimized on different KeyGNet inputs.
One KeyGNet is trained  
and evaluated in SISO mode
with the input of heuristically selected keypoints,
with
KGN optimization (FPS-KGN and BBox-KGN). 
An alternate KeyGNet is trained on segments, and estimates keypoints for all object categories in MIMO mode (KGN).
As shown in Table~\ref{tab:initPoints}, KeyGNet trained with object segment input data is slightly more accurate than those with heuristically defined keypoints as input.
In our subsequent experiments, we trained on segments, with one set of keypoints for all objects.


\section{Conclusion}
\label{sec:Conclusion}
In summary, we proposed KeyGNet to select a set of pre-defined dispersed keypoints with optimal similarly distributed votes for MIMO keypoint voting-based 6DoF PE.
The keypoints were selected by training a graph network to make the histograms of votes share similar distributions, so that the regression network can estimate votes more accurately.
Our keypoints improved the performance and training time on all seven BOP core datasets among all three SOTA methods tested, and reduced the SISO-MIMO degradation of these methods. 

\noindent {\bf Acknowledgements:} We thank Bluewrist Inc. and NSERC for their support of this work.


{\small
\bibliographystyle{ieee_fullname}
\bibliography{bib}
}

\end{document}


\title{Learning Better Keypoints for Multi-Object 6DoF Pose Estimation\\
Supplementary Material}

\author{Yangzheng Wu
, Michael Greenspan
\\ RCV Lab, Dept. of Electrical and Computer Engineering, Ingenuity Labs, \\ Queen's University, Kingston, Ontario, Canada \\{\tt \{y.wu, michael.greenspan\}@queensu.ca}
}
\maketitle

\section{Overview}
\begin{figure*}[ht]
    \centering
    \includegraphics[width=\textwidth]{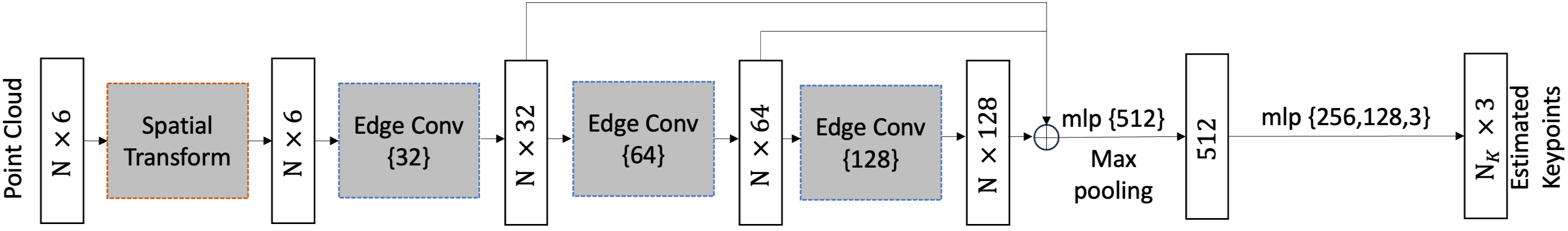}
    \caption{KeyGNet Network Structure. The network is based on the classification structure of edge-conv~\cite{wang2019dynamic}. The spatial transform block and edge-conv blocks are exactly the same as in the original setup. The output vector is reshaped to $N_K \times 3$ keypoints.
    \label{fig:edgeconv}}
\end{figure*}

We document here the network structure, some additional results, and one more ablation study. 
The network diagram is shown in Figure.~\ref{fig:edgeconv}. It is directly taken from the classification structure of edge-conv~\cite{wang2019dynamic} with a few changes to the number of intermediate channels and the shape of the output vector. 
The per category KeyGNet results of LMO and YCB-V datasets evaluated by ADD(S) and ADD(S) AUC in both SISO and MIMO modes on all three methods tested are shown in Table.~\ref{tab:lmo}, Table.~\ref{tab:YCBVideoFull_mimo}, and Table.~\ref{tab:YCBVideoFull_siso}.
KeyGNet keypoints improved the perfomance for all objects, in all datasets, among all three methods tested.
The BOP $AR$ (Average Recall) of Visible Surface Discrepancy ($AR_{VSD}$), Maximum Symmetry-Aware Surface Distance ($AR_{MSSD}$), Maximum Symmetry-Aware Projection Distance ($AR_{MSPD}$), and the overall average are reported in Table.~\ref{tab:bopFull}.
All these metrics are improved in all six core datasets when the KeyGNet keypoints are used.
Last but not least, the SISO-MIMO gap is reduced by using KeyGNet keypoints for all objects in YCB-V, as shown in Table.~\ref{tab:sisoVSmimo(ycb)}.

\section{Classical Distance Measure vs. KeyGNet}
Instead of training a network, keypoints can be selected by measuring the Wasserstein distance directlyon a collection of sets of keypoints. 
We conduct a test by comparing the trained KeyGNet with a classical RANSAC~\cite{fischler1981ransac} style algorithm. 
The collection of initial keypoint sets are selected either relatively randomly in a region centered at the bounding box's corners, or completely randomly in a sphere within the object reference frame of the CAD model. 
The Wasserstein distances and the dispersion scores are then calculated for each set of the keypoints. 
The algorithm repeats for $N$ times and the keypoints with the minimum Wasserstein distances and the maximum dispersion scores are selected.

\begin{table}[H]
\begin{center}
\begin{tabular}{@{}c|c|c|c@{}}
\hline
LMO&\multicolumn{2}{c|}{Random}&\multirow{2}{*}{KGN} \\\cline{2-3}
  Object   & BBox & Sphere   &   \\ \hline
ape         & 53.7 & \underline{55.2} & \textbf{65}   \\ 
can         & 80.8 & \underline{83.2} & \textbf{96.4}   \\ 
cat         & 44.1 & \underline{47.3} & \textbf{58}   \\ 
driller     & 70.6 & \underline{73.4} & \textbf{82.1}   \\ 
duck        & 42.1 & \underline{48.2} & \textbf{65.6}   \\ 
eggbox      & 70.1 & \underline{74.3} & \textbf{82.2  }   \\ 
glue        & 66.2 & \underline{67.3} & \textbf{75.1}   \\ 
holepuncher & 68.5 & \underline{72.5} & \textbf{81.2}   \\ \hline
average     &  62  & \underline{65.2} & \textbf{75.6}   \\\hline
\end{tabular}
\caption{
ADD(S) of RCVPose~\cite{wu2021vote} on LMO using keypoints selected randomly (BBox, Sphere) vs. with KeyGNet (KGN). The randomly selected keypoints  
use RANSAC to minimize the  Wasserstein distance measure.\label{tab:classicalvskgn}}
\end{center}
\end{table}

We test the keypoints using RCVPose on LMO and compare the ADD(S) metric with KeyGNet. The results are shown in Table.~\ref{tab:classicalvskgn}.
It can be seen that he keypoints selected with an initial location of bounding box corners are $3.2\%$ on average worse than those selected with completely random initial locations.
This is 
possibly due to the restrictions caused by the initial input locations of BBox corners.
The learned KeyGNet keypoints
have the best performance for all objects in LMO, boosting the ADD(S) by $13.6\%$ and $10.4\%$ compared to those
randomly selected.

{\small
\bibliographystyle{ieee_fullname}
\bibliography{bib}
}
\begin{table*}[]
\begin{center}
\begin{adjustbox}{max width=\textwidth}
\begin{tabular}{@{}c|c|c|c|c|c|c|c|c|c|c|c|c@{}}
\cline{2-13}
&\multicolumn{6}{|c|}{SISO} & \multicolumn{6}{|c}{MIMO}\\\hline
LMO          & \multicolumn{2}{c|}{PVNet}                  & \multicolumn{2}{c|}{PVN3D}                    & \multicolumn{2}{c}{RCVPose}    & \multicolumn{2}{|c|}{PVNet}                  & \multicolumn{2}{c|}{PVN3D}                    & \multicolumn{2}{c}{RCVPose}       \\\cline{2-13}
object & FPS & KGN & FPS & KGN & BBox & KGN& FPS & KGN & FPS & KGN & BBox & KGN\\\hline
ape            &  	15.8 &	21.2	 & 33.9 &	40.2	   & 61.3	& \textbf{65.4}	& 8.2	        & 20.9 &  25.7	       & 39.9	& 57.1	& \underline{65}          \\
can            &  	63.3 &	74.2	 & 88.6 &	\underline{93.7}	   & 93	    & \textbf{96.4}	& 51.1	        & 74.2 &  77.4	       & \underline{93.7}	& 90.1	& \textbf{96.4}\\
cat            &  	16.7 &	22.3	 & 39.1 &	49.2	   & 51.2	& \textbf{58.2}	& 9.6	        & 22.3 &  31.5	       & 48.9	& 45.5	& \underline{58} \\
driller        &  	65.7 &	76.6	 & 78.4 &	88.3	   & 78.8	& \underline{81.7}	& 57.5	        & 76.6 &  68.1	       & 88.3	& 77.5	& \textbf{82.1} \\
duck           &  	25.2 &	30.2	 & 41.9 &	47.6	   & 53.4	& \underline{65.2}	& 14.1	        & 29.9 &  32.2	       & 47.2	& 46.2	& \textbf{65.6} \\
eggbox         &  	50.2 &	57.8	 & 80.9 &	85.2	   & 82.3	& \underline{82}	& 38	        & 57.8 &  70.7	       & 85.2	& 80.2	& \textbf{82.2} \\
glue           &  	49.6 &	59.7	 & 68.1 &	77.2	   & 72.9	& \underline{74.9}	& 42.9	        & 59.7 &  58.3	       & 76.8	& 66.6	& \textbf{75.1} \\
holepuncher    &  	39.7 &	42.3	 & 74.7 &	82.3	   & 75.8	& \underline{81.2}	& 32.4	        & 42.3 &  67.5	       & \textbf{82.3}	& 73.7	& \underline{81.2} \\\hline
average        &    40.8 &	48 (+7.2)& 63.2 &	70.5 (+7.3)& 71.1	& 75.5 (+4.4) 	& 31.7 &  47.9 (\textbf{+16.2}) & 53.9  & 70.3	& 64.8	& 75.6 (\underline{+10.8}) \\\hline
\end{tabular}
\end{adjustbox}
\end{center}
\caption{LMO Results: The ADD(S) AUC comparison of three keypoint voting-based methods (PVNet, PVN3D, RCVPose) using initially defined keypoints (FPS, BBox) and optimized keypoints (KGN) generated by KeyGNet.\label{tab:lmo}}
\end{table*}

\begin{table*}[ht]
\begin{center}
\begin{tabular}{c|c|c|c|c|c|c}
\hline
YCB                  & \multicolumn{2}{c|}{PVNet} & \multicolumn{2}{c|}{PVN3D} & \multicolumn{2}{c}{RCVPose} \\ \cline{2-7}
     object                 & FPS & KGN & FPS & KGN & BBox & KGN \\ \hline
002\_master\_chef\_can   	    & 54.6	& 69.5 &	75.3 &	84.8  & 	\underline{92.1}  & 	\textbf{95.1} \\
003\_cracker\_box        	    & 66.2	& 78.8 &	87.0 &	\textbf{96.5}  & 	94.3  & 	\underline{96.4} \\
004\_sugar\_box          	    & 66.3	& 76.1 &	90.9 &	\underline{96.5}  & 	94.2  & 	\textbf{97.7} \\
005\_tomato\_soupcan     	    & 62.2	& 74.7 &	81.5 &	\underline{92.8}  & 	91.5  & 	\textbf{96.7} \\
006\_mustard\_bottle     	    & 67.6	& 82.4 &	89.3 &	\textbf{97.7}  & 	94.2  & 	\underline{96.9} \\
007Auna\_fish\_can       	    & 64.9	& 77.0 &	87.0 &	\underline{95.7}  & 	94.2  & 	\textbf{96.4} \\
008\_pudding\_box        	    & 76.6	& 85.2 &	90.3 &	\textbf{97.5}  & 	95.4  & 	\underline{97.0} \\
009\_gelatin\_box        	    & 71.4	& 88.8 &	90.6 &	\textbf{97.5}  & 	92.6  & 	\underline{97.4} \\
010\_potted\_meat\_can   	    & 69.5	& 84.1 &	82.5 &	\underline{93.7}  & 	88.1  & 	\textbf{93.9} \\
011\_banana              	    & 67.9	& 76.8 &	90.9 &	\underline{97.2}  & 	94.9  & 	\textbf{97.5} \\
019\_pitcher\_base       	    & 67.8	& 76.9 &	88.1 &	\underline{97.5}  & 	93.1  & 	\textbf{97.9} \\
021\_bleach\_cleanser    	    & 70.2	& 74.9 &	92.2 &	\underline{96.6}  & 	95.4  & 	\textbf{98.4} \\
024\_bowl$^{*}$          	    & 66.9	& 78.3 &	87.3 &	\underline{95.6}  & 	91.0  & 	\textbf{97.3} \\
025\_mug                 	    & 71.5	& 79.8 &	91.8 &	\underline{96.5}  & 	94.2  & 	\textbf{96.9} \\
035\_power\_drill        	    & 67.6	& 81.7 &	89.2 &	\underline{96.9}  & 	93.7  & 	\textbf{97.5} \\
036\_wood\_block$^{*}$   	    & 57.4	& 85.2 &	82.9 &	\underline{92.8}  & 	89.7  & 	\textbf{93.3} \\
037\_scissors            	    & 64.2	& 80.4 &	83.2 &	91.5  & 	\underline{92.3}  & 	\textbf{95.9} \\
040\_large\_marker       	    & 65.5	& 81.9 &	84.2 &	\underline{88.0}  & 	86.5  & 	\textbf{95.6} \\
051\_large\_clamp$^{*}$  	    & 55.7	& 66.5 &	84.4 &	88.5  & 	\underline{90.5}  & 	\textbf{97.6} \\
052\_extra\_large\_clamp$^{*}$ 	& 52.6	& 61.4 &	77.8 &	\underline{94.1}  &	    92.5  & 	\textbf{96.0} \\
061\_loam\_brick$^{*}$   	    & 60.9	& 79.6 &    89.4 &	\textbf{97.8}  & 	92.2  & 	\underline{97.2} \\\hline
average         	            & 65.1  & 78.1 &    86.5 &	\underline{94.6}  &  	92.5  & 	\textbf{96.6} \\ \hline
\end{tabular}
\caption{YCB-V Results. The ADD(S) AUC comparison of three keypoint voting-based methods (PVNet, PVN3D, RCVPose) in MIMO mode using initially defined keypoints (FPS, BBox) and optimized keypoints (KGN) generated by KeyGNet.
\label{tab:YCBVideoFull_mimo}
}
\end{center}
\end{table*}

\begin{table*}[ht]
\begin{center}
\begin{tabular}{c|c|c|c|c|c|c}
\hline
YCB                  & \multicolumn{2}{c|}{PVNet} & \multicolumn{2}{c|}{PVN3D} & \multicolumn{2}{c}{RCVPose} \\ \cline{2-7}
     object                 & FPS & KGN & FPS & KGN & BBox & KGN \\ \hline
002\_master\_chef\_can   & 60.2 & 70  & 79.3 & 85.2  &  \underline{94.7} &  \textbf{96.2}  \\ 
003\_cracker\_box        & 70.7 & 79.4  & 91.5 &  \underline{96.7} &  96.4 & \textbf{97.4}   \\ 
004\_sugar\_box          & 73.2 & 76.6  & 96.9 &  97.3 &  \underline{97.6} & \textbf{98.7}   \\ 
005\_tomato\_soupcan     & 67.7 &  75.1 & 89.0 & 93.2  &  \underline{95.4} &   \textbf{97.6} \\ 
006\_mustard\_bottle     & 76.5 & 83  & \underline{97.9} & \textbf{98.2}  & 97.7  &  \textbf{98.2}  \\ 
007Auna\_fish\_can       & 71.3 &  77.2 & 90.7 & 96.3  &  \underline{96.7} &  \textbf{97.4}  \\ 
008\_pudding\_box        & 80.1 & 85.4  & 97.1 & \textbf{98.1}  &  97.4 &   \underline{97.9} \\ 
009\_gelatin\_box        & 81.2 & 89.1  & \textbf{98.3} & \textbf{98.3}  &  \underline{97.9} &   \textbf{98.3} \\ 
010\_potted\_meat\_can   & 76.9 &  84.6 & 87.9 &  \underline{94.2} & 92.6  &  \textbf{95.3}  \\ 
011\_banana              & 73.2 & 77.6  & 96.0 &  \underline{97.6} & 97.2  &  \textbf{98.4} \\ 
019\_pitcher\_base       & 74.3 &  77.4 & 96.9 & \underline{98.0}  &  96.7 &  \textbf{99.2}  \\ 
021\_bleach\_cleanser    & 70.9 &  75.4 & 95.9 &  97.3 & \underline{98.4}  &  \textbf{99.3} \\ 
024\_bowl$^{*}$          & 69.7 & 79  & 92.8 & \underline{96.4}  & 95.3  &   \textbf{98.2} \\ 
025\_mug                 & 75.3 & 80.6  & 96.0 & \underline{97.1}  &  \underline{97.1} &  \textbf{98}  \\ 
035\_power\_drill        & 74.3 & 82  & 95.7 &  \underline{97.2} & 96.9  &  \textbf{98.3}  \\ 
036\_wood\_block$^{*}$   & 70.2 & 85.8  & 91.1 & \underline{93.2}  &  90.7 &  \textbf{94.3}  \\ 
037\_scissors            & 66.4 &  81 & 87.2 & 92.1  & \underline{94.9}  &  \textbf{97.2}  \\ 
040\_large\_marker       & 67.3 &  82.4 & 91.6 &  \underline{94.3} &  93.2 &  \textbf{96.3}  \\ 
051\_large\_clamp$^{*}$  & 66.2 & 72.2  & 95.6 &  \underline{96.2} & \underline{96.2}  &  \textbf{98.3}  \\ 
052\_extra\_large\_clamp$^{*}$ & 63.4 &  66.9 & 90.5 &  94.7 &  \underline{95.1} &  \textbf{97.2}  \\ 
061\_loam\_brick$^{*}$   & 70.2 &  80.3 & \underline{98.2} & \textbf{98.4}  &  96.6 &  \underline{98.2}  \\ \hline
average         & 73.4 & 79.1  & 92.3 & 95.7  &  \underline{95.9} &  \textbf{97.6} \\ \hline
\end{tabular}
\caption{YCB-V Results. The ADD(S) AUC comparison of three keypoint voting-based methods (PVNet, PVN3D, RCVPose) in SISO mode using initially defined keypoints (FPS, BBox) and optimized keypoints (KGN) generated by KeyGNet.
\label{tab:YCBVideoFull_siso}
}
\end{center}
\end{table*}
\begin{table*}[ht]
\begin{center}
\begin{tabular}{c|c|c|c|c|c|c|c}
\hline
\multirow{2}{*}{Metric} &   \multirow{2}{*}{Dataset}              & \multicolumn{2}{c|}{PVNet} & \multicolumn{2}{c|}{PVN3D} & \multicolumn{2}{c}{RCVPose} \\ \cline{3-8}
            &          & FPS & KGN & FPS & KGN & BBox & KGN \\ \hline
\multirow{7}{*}{$AR_{VSD}$}	  	&  LMO      &  	48.2&	52.4&	70.6&	\underline{72.7}&	72.5&	\textbf{76.9}	  \\ \cline{2-8}
	    					  	&  YCB-V    &  	78.2&	82.7&	76.9&	83.6&	\underline{84.4}&	\textbf{88.3}	  \\ \cline{2-8}
	    					  	&  TLESS    &  	65.7&	67.2&	68.3&	\underline{72.2}&	70.8&	\textbf{75.3}	  \\ \cline{2-8}
	    					  	&  TUDL     & 	90.5&	92.5&	87.3&	91.9&	\underline{98.0}&	\textbf{99.4}	  \\ \cline{2-8}
	    					  	&  IC-BIN   &   70.6&	72.6&	67.2&	73.9&	\underline{74.1}&	\textbf{80.4}	  \\ \cline{2-8}
	    					  	&  ITODO    &  	42.4&	44.7&	43.2&	47.7&	\underline{50.7}&	\textbf{50.8}	  \\ \cline{2-8}
	    					  	&  HB      	&   77.1&	78.7&	78.4&	\underline{84.4}&	82.5&	\textbf{85.9}	  \\ \cline{1-8}
\multirow{7}{*}{$AR_{MSSD}$}  	&  LMO      &  	66.4&	70.2&	62.5&	69.3&	\underline{73.4}&	\textbf{73.6}	  \\ \cline{2-8}
	    					  	&  YCB-V    &  	77.3&	79.8&	79.9&	82.9&	\underline{86.3}&	\textbf{89.6}	  \\ \cline{2-8}
	    					  	&  TLESS    &  	70.2&	72.0&	64.3&	66.5&	\underline{72.3}&	\textbf{73.3}	  \\ \cline{2-8}
	    					  	&  TUDL     & 	90.6&	91.8&	91.9&	93.1&	\underline{97.5}&	\textbf{98.5}	  \\ \cline{2-8}
	    					  	&  IC-BIN   &   69.0&	71.5&	72.1&	\textbf{78.4}&	\underline{73.9}&	73.8	  \\ \cline{2-8}
	    					  	&  ITODO    &  	51.2&	51.3&	52.6&	\underline{58.3}&	57.2&	\textbf{64.2}	  \\ \cline{2-8}
	    					  	&  HB      	&   85.5&	\underline{90.2}&	85.3&	88.2&	89.0&	\textbf{90.4}	  \\ \cline{1-8}
\multirow{7}{*}{$AR_{MSPD}$}  	&  LMO      &  	69.2&	71.7&	66.0&	67.7&	\underline{74.9}&	\textbf{79.7}	  \\ \cline{2-8}
	    					  	&  YCB-V    &  	76.7&	77.0&	76.5&	81.2&	\underline{84.9}&	\textbf{88.1}	  \\ \cline{2-8}
	    					  	&  TLESS    &  	67.3&	71.1&	69.3&	\underline{72.1}&	71.5&	\textbf{77.7}	  \\ \cline{2-8}
	    					  	&  TUDL     & 	93.7&	94.9&	91.2&	95.1&	\underline{97.8}&	\textbf{98.6}	  \\ \cline{2-8}
	    					  	&  IC-BIN   &   72.3&	75.4&	71.8&	\textbf{77.6}&	73.9&	\underline{75.6}	  \\ \cline{2-8}
	    					  	&  ITODO    &  	50.3&	52.3&	52.7&	55.6&	\underline{56.2}&	\textbf{59.4}	  \\ \cline{2-8}
	    					  	&  HB      	&   84.8&	86.0&	84.7&	90.2&	\underline{90.5}&	\textbf{93.0}	  \\ \cline{1-8}
\multirow{7}{*}{$AR_{average}$}	&  LMO      &  	61.3&	64.8&	66.4&	69.9&	\underline{73.6}&	\textbf{76.7}	  \\ \cline{2-8}
	    						&  YCB-V    &  	77.4&	79.8&	77.8&	82.6&	\underline{85.2}&	\textbf{88.7}	  \\ \cline{2-8}
	    						&  TLESS    &  	67.7&	70.1&	67.3&	70.3&	\underline{71.5}&	\textbf{75.4}	  \\ \cline{2-8}
	    						&  TUDL     & 	91.6&	93.1&	90.1&	93.4&	\underline{97.8}&	\textbf{98.8}	  \\ \cline{2-8}
	    						&  IC-BIN   &   70.6&	73.2&	70.4&	\textbf{76.6}&	\underline{74.0}&	\textbf{76.6}	  \\ \cline{2-8}
	    						&  ITODO    &  	48.0&	49.4&	49.5&	53.9&	\underline{54.7}&	\textbf{58.1}	  \\ \cline{2-8}
	    						&  HB      	&   82.5&	85.0&	82.8&	\underline{87.6}&	87.3&	\textbf{89.7}	  \\ \cline{1-8}
\multicolumn{2}{c|}{Overall Average}        &   71.3& 73.6&72.0&76.3&\underline{77} &\textbf{80.6}     \\\hline
\end{tabular}
\caption{BOP Core Dataset Results. The Average Recall ($AR$) of Visible Surface Discrepancy ($AR_{VSD}$), Maximum Symmetry-Aware Surface Distance ($AR_{MSSD}$), Maximum Symmetry-Aware Projection Distance ($AR_{MSPD}$), and the overall average for all six BOP core datasets are reported for three methods by using keypoints selected by the original method (FPS/BBox) and KeyGNet (KGN).
\label{tab:bopFull}
}
\end{center}
\end{table*}

\begin{table*}[ht]
\begin{center}
\begin{tabular}{c|c|c|c|c|c|c}
\hline
YCB-V                  & \multicolumn{2}{c|}{PVNet} & \multicolumn{2}{c|}{PVN3D} & \multicolumn{2}{c}{RCVPose} \\ \cline{2-7}
     object                 & FPS & KGN & FPS & KGN & BBox & KGN \\ \hline
002\_master\_chef\_can   	    & -5.6	& \underline{-0.5}	& -4.0	 &\textbf{-0.4}	&-2.6	&-1.1 \\ 
003\_cracker\_box        	    & -4.5	& \underline{-0.6}	& -4.5	 &\textbf{-0.2}	&-2.1	&-1.0 \\ 
004\_sugar\_box          	    & -6.9	& \textbf{-0.5}	& -6.0	 &\underline{-0.8}	&-3.4	&-1.0 \\ 
005\_tomato\_soupcan     	    & -5.5	& \textbf{-0.4}	& -7.5	 &\textbf{-0.4}	&-3.9	&\underline{-0.9} \\ 
006\_mustard\_bottle     	    & -8.9	& \underline{-0.6}	& -8.6	 &\textbf{-0.5}	&-3.5	&-1.3 \\ 
007Auna\_fish\_can       	    & -6.4	& \textbf{-0.2}	& -3.7	 &\underline{-0.6}	&-2.5	&-1.0 \\ 
008\_pudding\_box        	    & -3.5	& \textbf{-0.2}	& -6.8	 &\underline{-0.6}	&-2.0	&-0.9 \\ 
009\_gelatin\_box        	    & -9.8	& \textbf{-0.3}	& -7.7	 &\underline{-0.8}	&-5.3	&-0.9 \\ 
010\_potted\_meat\_can   	    & -7.4	& \textbf{-0.5}	& -5.4	 &\textbf{-0.5}	&-4.5	&\underline{-1.4} \\ 
011\_banana              	    & -5.3	& \underline{-0.8}	& -5.1	 &\textbf{-0.4}	&-2.3	&-0.9 \\ 
019\_pitcher\_base       	    & -6.5	& \textbf{-0.5}	& -8.8	 &\textbf{-0.5}	&-3.6	&\underline{-1.3} \\ 
021\_bleach\_cleanser    	    & \underline{-0.7}	& \textbf{-0.5}	& -3.7	 &\underline{-0.7}	&-3.0	&-0.9 \\ 
024\_bowl$^{*}$          	    & -2.8	& \textbf{-0.7}	& -5.5	 &\underline{-0.8}	&-4.3	&-0.9 \\ 
025\_mug                 	    & -3.8	& \underline{-0.8}	& -4.2	 &\textbf{-0.6}	&-2.9	&-1.1 \\ 
035\_power\_drill        	    & -6.7	& \textbf{-0.3}	& -6.5	 &\textbf{-0.3}	&-3.2	&\underline{-0.8} \\ 
036\_wood\_block$^{*}$   	    & -12.8	& \underline{-0.6}	& -8.2	 &\textbf{-0.4}	&-1.0	&-1.0 \\ 
037\_scissors            	    & -2.2	& \textbf{-0.6}	& -4.0	 &\textbf{-0.6}	&-2.6	&\underline{-1.3} \\ 
040\_large\_marker       	    & -1.8	& \textbf{-0.5}	& -7.4	 &-6.3	&-6.7	&\underline{-0.7} \\ 
051\_large\_clamp$^{*}$  	    & -10.5	& \underline{-5.7}	& -11.2  &-7.7  &\underline{-5.7}   &\textbf{-0.7} \\ 
052\_extra\_large\_clamp$^{*}$ 	& -10.8	& -5.5	& -12.7  &\textbf{-0.6}  &-2.6   &\underline{-1.2} \\ 
061\_loam\_brick$^{*}$   	    & -9.3	& \underline{-0.7}	& -8.8	 &\textbf{-0.6}	&-4.4	&-1.0 \\ \hline
average         	            & -6.3	& \textbf{-1.0}	& -6.7	 &\underline{-1.2}	&-3.4	&\textbf{-1.0} \\ \hline
\end{tabular}
\caption{SISO-MIMO performance gap on YCB-V. The change in ADD(S) when converting from SISO to MIMO, for keypoints sampled heuristically (FPS or BBox) and KeyGNet (KGN). There is a relatively small change in ADD(S) AUC when the PE network is trained simultaneously on multiple objects
using 
KGN keypoints.
\label{tab:sisoVSmimo(ycb)}
}
\end{center}
\end{table*}

